\documentclass[letterpaper, 10 pt, conference]{ieeeconf}  % Comment this line out if you need a4paper

\IEEEoverridecommandlockouts                              % This command is only needed if 
\usepackage[utf8]{inputenc}
\usepackage{graphicx}
\usepackage{hyperref}
\usepackage{gensymb}
\usepackage[T1]{fontenc}% Needed for \textquotedbl
\usepackage{soul}

\newcommand\rurl[1]{%
  \href{http://#1}{\nolinkurl{#1}}%
}
\usepackage{listings}% http://ctan.org/pkg/listings
\newif\ificra
\icratrue
\icrafalse

\usepackage[noadjust]{cite}

\usepackage{cleveref}
\crefname{table}{Table}{Tables}
\crefname{figure}{Figure}{Figures}
\crefname{section}{Section}{Sections}

\usepackage[binary-units]{siunitx}
\DeclareSIUnit\uint{uint8}
\DeclareSIUnit\pixel{px}

\usepackage[acronym]{glossaries}
\setglossarystyle{long}

\makeglossaries

\newacronym{fmcw}{FMCW}{Frequency-Modulated Continuous-Wave}
\newacronym{ori}{ORI}{Oxford Robotics Institute}
\newacronym{vo}{VO}{Visual Odometry}
\newacronym{gps}{GPS}{Global Positioning System}
\newacronym{sdk}{SDK}{Software Development Kit}
\newacronym{dl}{DL}{Deep Learning}
\newacronym{slerp}{SLERP}{spherical linear interpolation}

\usepackage{xcolor}
\definecolor{CommentBarnes}{rgb}{1,0,0}

\definecolor{CommentGadd}{rgb}{0,0.8,0}

\definecolor{CommentMurcutt}{rgb}{0,0,1}

\definecolor{CommentNewman}{rgb}{1,0,0.5}

\definecolor{CommentPosner}{rgb}{1,0.5,0}

\newcolumntype{C}[1]{>{\centering\let\newline\\\arraybackslash\hspace{0pt}}m{#1}}

\newcommand{\doctitle}
{The Oxford Radar RobotCar Dataset:\\ A Radar Extension to the Oxford RobotCar Dataset }
\newcommand{\docsubtitle}{}

\title{\LARGE \bf\doctitle\docsubtitle} 
\author{Dan Barnes, Matthew Gadd, Paul Murcutt, Paul Newman and Ingmar Posner
\thanks{Authors are from the Oxford Robotics Institute, University of Oxford, UK.
{\tt\small \{dbarnes,mattgadd,pmurcutt,pnewman,ingmar\} @robots.ox.ac.uk}}
}

\begin{document}

% disable ruling out of duplicate author lists in bibliography
\bstctlcite{IEEEexample:BSTcontrol}

\maketitle
\thispagestyle{empty}
\pagestyle{empty}

%%%%%%%%%%%%%%%%%%%%%%%%%%%%%%%%%%%%%%%%%%%%%%%%%%%%%%%%%%%%%%%%%%%%%%%%%%%%%%%%
%%% ABSTRACT
%%%%%%%%%%%%%%%%%%%%%%%%%%%%%%%%%%%%%%%%%%%%%%%%%%%%%%%%%%%%%%%%%%%%%%%%%%%%%%%%
\begin{abstract}
In this paper we present \emph{The Oxford Radar RobotCar Dataset}, a new dataset for researching scene understanding using Millimetre-Wave FMCW scanning radar data.
The target application is autonomous vehicles where this modality is robust to environmental conditions such as fog, rain, snow, or lens flare, which typically challenge other sensor modalities such as vision and LIDAR.

The data were gathered in January 2019 over thirty-two traversals of a central Oxford route spanning a total of \SI{280}{\kilo\metre} of urban driving. It encompasses a variety of weather, traffic, and lighting conditions. This \SI{4.7}{\tera\byte} dataset consists of over \SI{240000}{} scans from a Navtech CTS350-X radar and \SI{2.4}{} million scans from two Velodyne HDL-32E 3D LIDARs; along with six cameras, two 2D LIDARs, and a GPS/INS receiver. In addition we release ground truth optimised radar odometry to provide an additional impetus to research in this domain. The full dataset is available for download at: \\ \rurl{ori.ox.ac.uk/datasets/radar-robotcar-dataset}
\end{abstract}

%%%%%%%%%%%%%%%%%%%%%%%%%%%%%%%%%%%%%%%%%%%%%%%%%%%%%%%%%%%%%%%%%%%%%%%%%%%%%%%%
%%% INTRODUCTION
%%%%%%%%%%%%%%%%%%%%%%%%%%%%%%%%%%%%%%%%%%%%%%%%%%%%%%%%%%%%%%%%%%%%%%%%%%%%%%%%
\section{Introduction}

While many of the challenges in urban autonomy have been met successfully with lasers and cameras, radar offers the field of robotics an alternative modality for robust sensing.
The \gls{fmcw} class of radar provides a \mbox{\SI{360}{\degree}-view} of the scene and is capable of detecting targets at ranges far exceeding those of automotive 3D LIDAR.
These advantages are particularly valuable to autonomous vehicles which need to see further if they are to travel safely at higher speeds or to operate in wide open spaces where there is a dearth of distinct features.
Moreover, these vehicles must function reliably in unstructured environments and require a sensor such as radar that thrives in all conditions -- rain, snow, dust, fog, or direct sunlight.

This dataset builds upon the \textit{Oxford RobotCar Dataset}~\cite{RobotcarDatasetIJRR}, one of the the largest available datasets for autonomous driving research.
The original dataset release consisted of over \SI{20}{\tera\byte} of vehicle-mounted monocular and stereo imagery, 2D and 3D LIDAR, as well as inertial and GPS data collected over a year of driving in Oxford, UK.
More than \SI{100}{} traversals of a \SI{10}{\kilo\metre} route were performed over this period to capture scene variation over a range of timescales, from the \SI{24}{\hour} day/night illumination cycle to long-term seasonal variations.
As a valuable resource for self-driving research, the vehicle software and mechatronics have been maintained since the original dataset was gathered and released: now configured with a millimetre-wave radar and two additional 3D LIDARs.
The current appearance of the vehicle with these additional sensors can be seen in \cref{fig:robotcar-platform}.
Along with the raw sensor recordings from all sensors, we provide an updated set of calibrations, ground truth trajectory for the radar sensor as well as MATLAB and Python development tools for utilising the data. 

By sharing this large-scale radar dataset with researchers we aim to accelerate research into this promising modality for mobile robotics and autonomous vehicles of the future.

\begin{figure}
    \centering
    \includegraphics[width=\columnwidth]{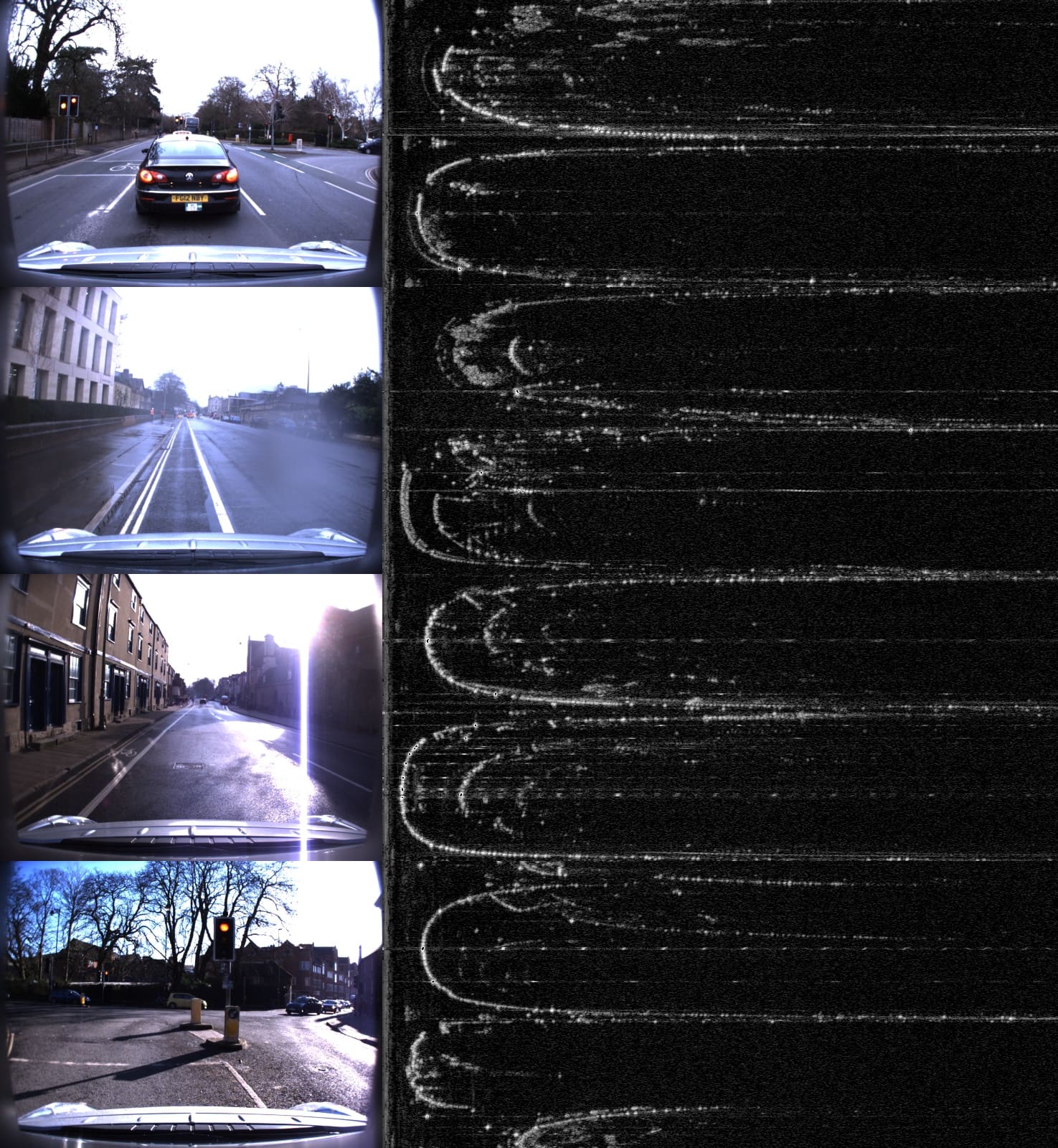}
    \caption{
    The Oxford Radar RobotCar Dataset for complex and robust scene understanding with Millimetre-Wave \gls{fmcw} scanning radar data. 
    We collected \SI{32}{} traversals of a central Oxford route with the Oxford RobotCar platform during the month of January, 2019.
    Despite weather conditions such as rain, direct sunlight, and fog which are challenging for traditional modalities such as vision (left), radar (right) holds the promise of consistent sensor observations for mapping, localisation, and scene understanding.
    Sample pairs are taken from different locations of the driven route. 
    } 
\label{fig:hero-figure}
\end{figure}

%%%%%%%%%%%%%%%%%%%%%%%%%%%%%%%%%%%%%%%%%%%%%%%%%%%%%%%%%%%%%%%%%%%%%%%%%%%%%%%%
%%% RELATED WORK
%%%%%%%%%%%%%%%%%%%%%%%%%%%%%%%%%%%%%%%%%%%%%%%%%%%%%%%%%%%%%%%%%%%%%%%%%%%%%%%%
\section{Related Work}

A number of LIDAR- and vision-based autonomous driving datasets, such as~\cite{pandey2011ford,geiger2013vision,blanco2014malaga,cordts2016cityscapes,yu2018bdd100k,nuscenes2019,lyft2019}, are available to the community and were primarily collected in order to develop competencies in these modalities.
This dataset release is meant to advocate the increased exploitation of \gls{fmcw} radar for vehicle autonomy. We therefore present radar data \emph{alongside} the camera and LIDAR data typically appearing these datasets with the goal of replicating and advancing these competencies with this promising sensor modality.

Similar radar sensors have been used in a variety of domains for mapping, navigation, and perception \cite{callmer2011radar,reina2011radar, adams2012robotic}.
Some publications using similar, if not identical, \gls{fmcw} radar for state estimation prior to the release of this dataset include~\cite{vivet2013localization,schuster2016landmark,2018ICRA_cen,2019ICRA_cen,2019ICRA_aldera}.
To this end, \cref{sec:gt-radar-odometry} discusses the optimised ground truth radar odometry data released as part of this dataset to help further research in this area.

The Navtech radar dataset presented in~\cite{2019ICRA_park} is concurrent to this release.
Although significantly smaller in size than our release, 
the comparable setups should provide a great opportunity for cross-validating approaches between datasets in different geographical locations.
The \textit{Marulan datasets} presented in~\cite{peynot2010marulan} also use \gls{fmcw} radar, but only configured to a maximum range of \SI{40}{\metre}.
Additionally, while these datasets are collected under variable conditions, they represent fairly static outdoor scenes that are not representative of urban driving.

%%%%%%%%%%%%%%%%%%%%%%%%%%%%%%%%%%%%%%%%%%%%%%%%%%%%%%%%%%%%%%%%%%%%%%%%%%%%%%%%
%%% THE RADAR ROBOTCAR PLATFORM
%%%%%%%%%%%%%%%%%%%%%%%%%%%%%%%%%%%%%%%%%%%%%%%%%%%%%%%%%%%%%%%%%%%%%%%%%%%%%%%%
\section{The Radar RobotCar Platform}

The dataset was collected using the Oxford RobotCar platform as in~\cite{RobotcarDatasetIJRR}, an autonomous-capable Nissan LEAF, illustrated with sensor layout in \cref{fig:robotcar-platform}.
For this release, the RobotCar was equipped with the following sensors which were not in the original release:

\begin{figure}
    \centering
    \includegraphics[width=\linewidth]{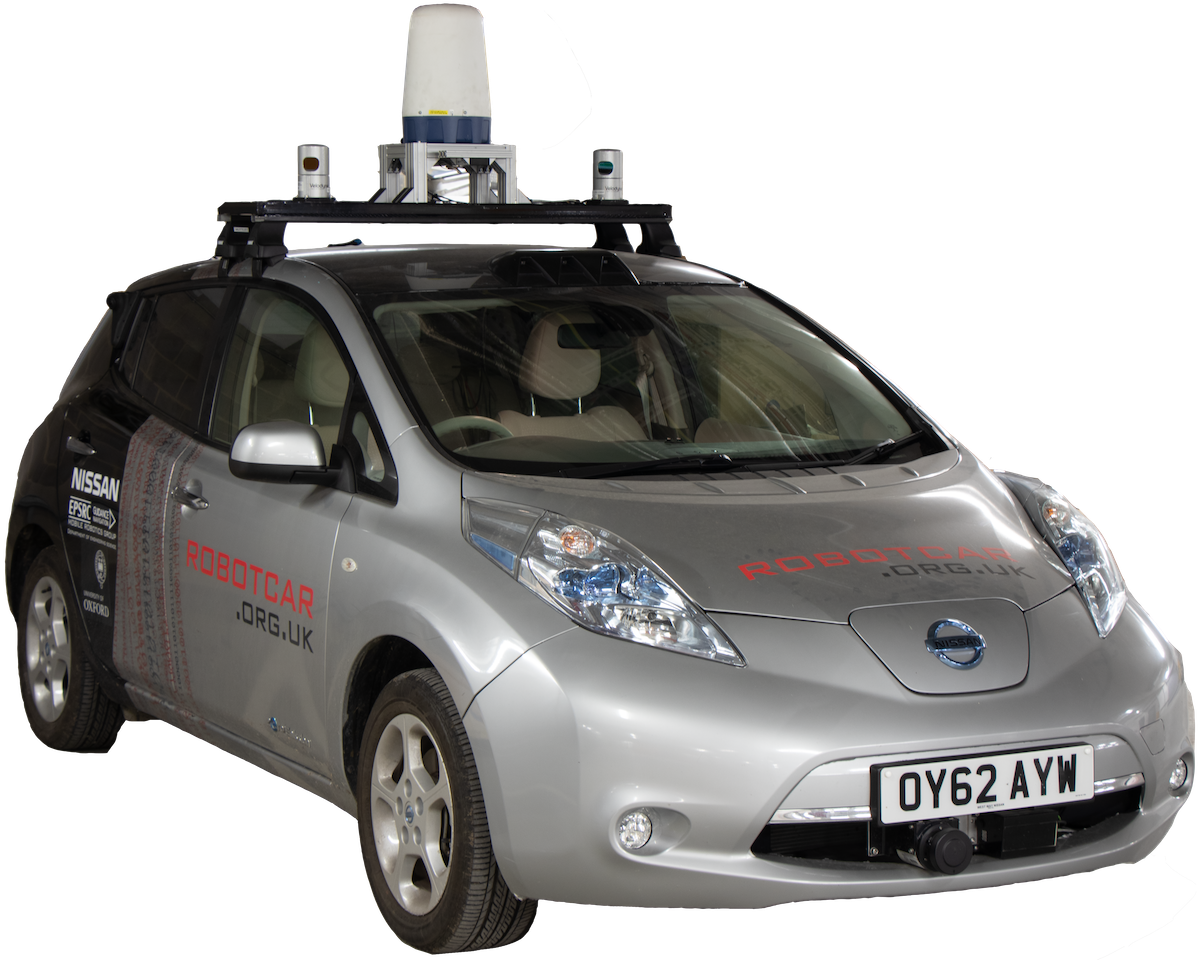}
    \includegraphics[width=\linewidth]{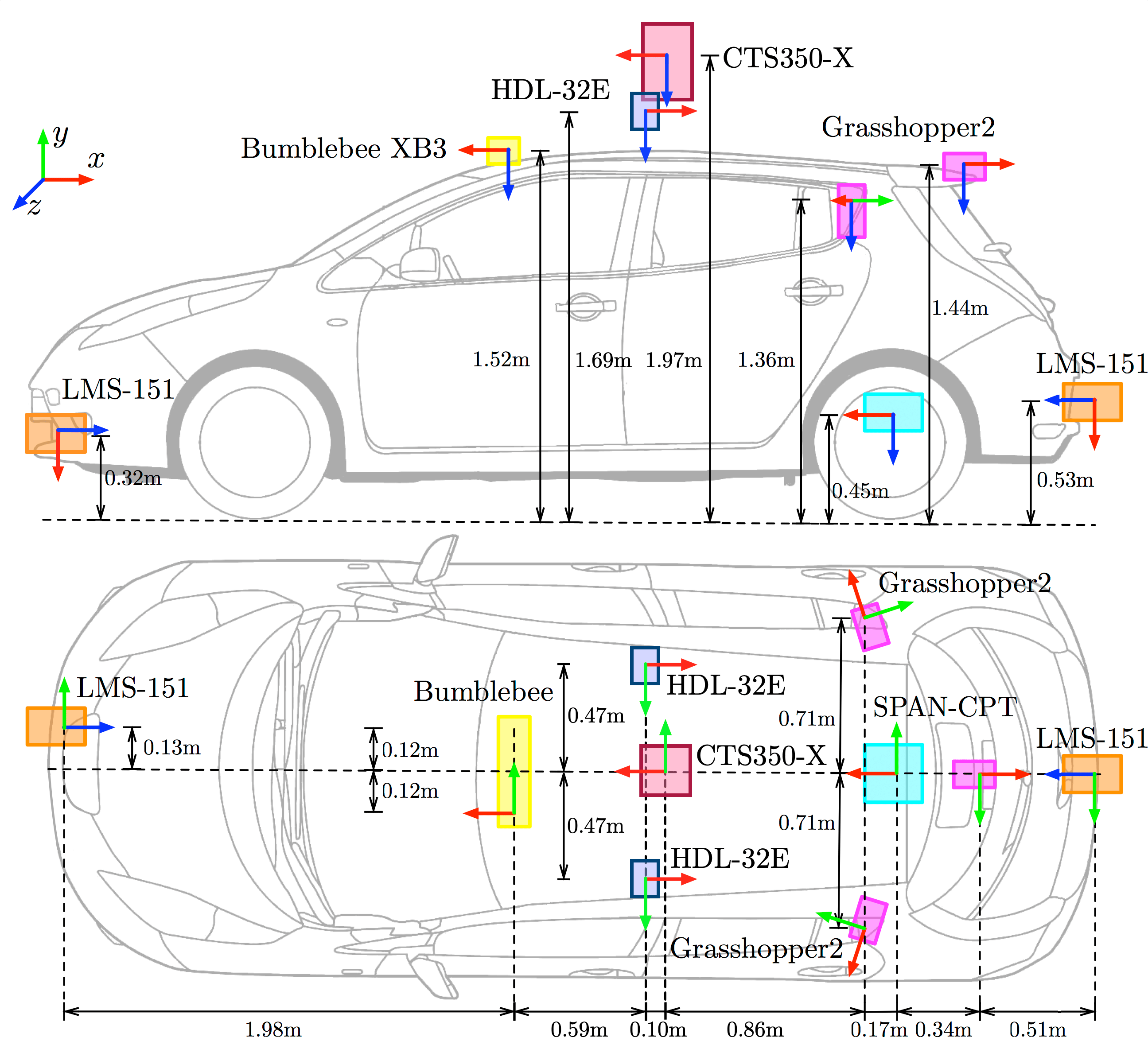}
    \caption{The Radar RobotCar platform (top) and sensor location diagram (bottom) with the Navtech CTS350-X radar mounted in the centre.
    Coordinate frames show the origin and direction of each sensor mounted on the vehicle with the convention: $x$-forward (red), $y$-right (green), $z$-down (blue).
    Measurements shown are approximate; the development tools include exact $SE(3)$ extrinsic calibrations for all sensors.  
    }
    \label{fig:robotcar-platform}
\end{figure}

\begin{itemize}
    \item 1 x Navtech CTS350-X Millimetre-Wave \gls{fmcw} radar, \SI{4}{\hertz}, \SI{400}{} measurements per rotation, \SI{163}{\metre} range, \SI{4.38}{\centi\metre} range resolution, \SI{1.8}{\degree}~beamwidth 
    \item 2 x Velodyne HDL-32E 3D LIDAR, \SI{360}{\degree}~HFoV, \SI{41.3}{\degree}~VFoV, \SI{32}{} planes, \SI{20}{\hertz}, \SI{100}{\metre} range, \SI{2}{\centi\metre} range resolution
\end{itemize}
In addition to the original sensors as in~\cite{RobotcarDatasetIJRR}:
\begin{itemize}
    \item 1 x Point Grey Bumblebee XB3 (BBX3-13S2C-38) trinocular stereo camera, \SI{1280}{}$\times$\SI{960}{}$\times$\SI{3}{}, \SI{16}{\hertz}, \SI{1/3}{}\textquotedbl~Sony ICX445 CCD, global shutter, \SI{3.8}{\milli\metre} lens, \SI{66}{\degree}~HFoV, \SI{12/24}{\centi\metre} baseline
    \item 3 x Point Grey Grasshopper2 (GS2-FW-14S5C-C) monocular camera, \SI{1024}{}$\times$\SI{1024}{}, \SI{11.1}{\hertz}, \SI{2/3}{}\textquotedbl~Sony ICX285 CCD, global shutter, \SI{2.67}{\milli\metre} fisheye lens (Sunex DSL315B-650-F2.3), \SI{180}{\degree}~HFoV
    \item 2 x SICK LMS-151 2D LIDAR, \SI{270}{\degree}~FoV, \SI{50}{\hertz}, \SI{50}{\metre} range, \SI{0.5}{\degree}~resolution
    \item 1 x NovAtel SPAN-CPT ALIGN inertial and GPS navigation system, \SI{6}{} axis, \SI{50}{\hertz}, GPS/GLONASS, dual antenna
\end{itemize}

As the main focus of this release, the \mbox{Navtech CTS350-X} radar was mounted at the centre of the vehicle aligned to the vehicle axes.
We used a pair of Velodyne HDL-32E 3D LIDARs instead of the LD-MRS 3D LIDAR used in~\cite{RobotcarDatasetIJRR} for drastically improved 3D scene understanding.
In addition to providing twice the range and intensity returns, the Velodynes provide a full \SI{360}{\degree}~HFoV with \SI{41.3}{\degree}~VFoV for full coverage around the vehicle.

Sensor drivers for both the Navtech CTS350-X and Velodyne HDL-32E devices were developed internally to provide accurate synchronisation and timestamping with the other sensors..
For further details on sensors from the original release, compute specifications, and data logging procedures please consult the original dataset paper~\cite{RobotcarDatasetIJRR}. 

\begin{figure*}
    \centering
    \includegraphics[width=\textwidth]{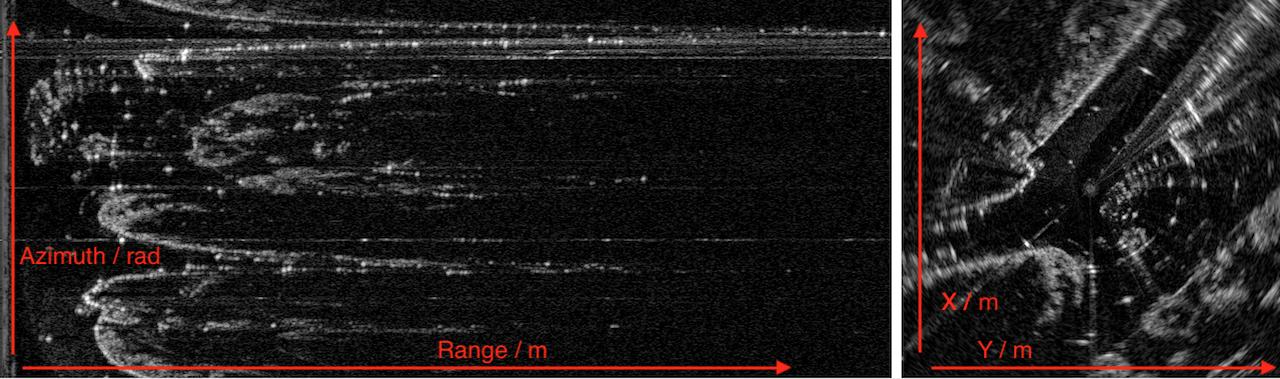}
    \caption{
    Example sensor data from the Navtech CTS350-X radar.
    Raw radar power power returns in polar form (left) for a full sweep of $0\rightarrow2\pi$ over a range of $0\rightarrow163\text{m}$ and the corresponding scan in Cartesian form (right), with the vehicle in the center and axes from $\text{-}50\text{m}\rightarrow50\text{m}$.
    Tools required to parse the data and perform the polar-to-Cartesian conversion are provided in the \gls{sdk} discussed in \cref{sec:sdk}.
    }
    \label{fig:radar-data}
\end{figure*}

%%%%%%%%%%%%%%%%%%%%%%%%%%%%%%%%%%%%%%%%%%%%%%%%%%%%%%%%%%%%%%%%%%%%%%%%%%%%%%%%
%%% RADAR DATA
%%%%%%%%%%%%%%%%%%%%%%%%%%%%%%%%%%%%%%%%%%%%%%%%%%%%%%%%%%%%%%%%%%%%%%%%%%%%%%%%
\section{Radar Data}

The Navtech CTS350-X is a \gls{fmcw} scanning radar without Doppler information, configured to return \SI{3768}{} power readings at a range resolution of \SI{4.38}{\centi\metre} across \SI{400} azimuths at a frequency of \SI{4}{\hertz} (corresponding to a maximum range of \SI{163}{\metre} and \SI{0.9}{\degree}~azimuth resolution). 
Other configurations of the Navtech CTS350-X are able to provide range in excess of \SI{650}{\metre} or higher rotation frequencies.
However, for this dataset shorter range, high resolution data was deemed most useful in urban scenarios where straight line distances over \SI{163}{\metre} are rare.

This type of radar rotates about its vertical axis while continuously transmitting and receiving frequency-modulated radio waves similar to a spinning LIDAR.
The frequency shift between the transmitted and received waves is used to compute the range of an object, and the received power is a function of the object’s reflectivity, size, shape, and orientation relative to the receiver. 
One full rotation and its 2D power data can be represented by a matrix in which each row corresponds to an azimuth and each column to a range, as shown in \cref{fig:radar-data}, where the intensity represents the highest power reflection within a range bin.

The radar operates at frequencies of \SI{76}{\giga\hertz} to \SI{77}{\giga\hertz}, ensuring consistent measurements through harsh local conditions such as dust, rain, and snow.
The main beam spread is \SI{1.8}{\degree} between \SI{-3}{\decibel} points horizontally and vertically; with an additional cosec squared fill-in beam pattern up to \SI{40}{\degree} below the horizontal which permits detection of objects beneath the main beam.

%%%%%%%%%%%%%%%%%%%%%%%%%%%%%%%%%%%%%%%%%%%%%%%%%%%%%%%%%%%%%%%%%%%%%%%%%%%%%%%%
%%% DATA COLLECTION
%%%%%%%%%%%%%%%%%%%%%%%%%%%%%%%%%%%%%%%%%%%%%%%%%%%%%%%%%%%%%%%%%%%%%%%%%%%%%%%%
\section{Data Collection}

\ificra
\else
\begin{figure*}
    \centering
    \includegraphics[width=\linewidth,trim={0 0 0 14.13cm},clip]{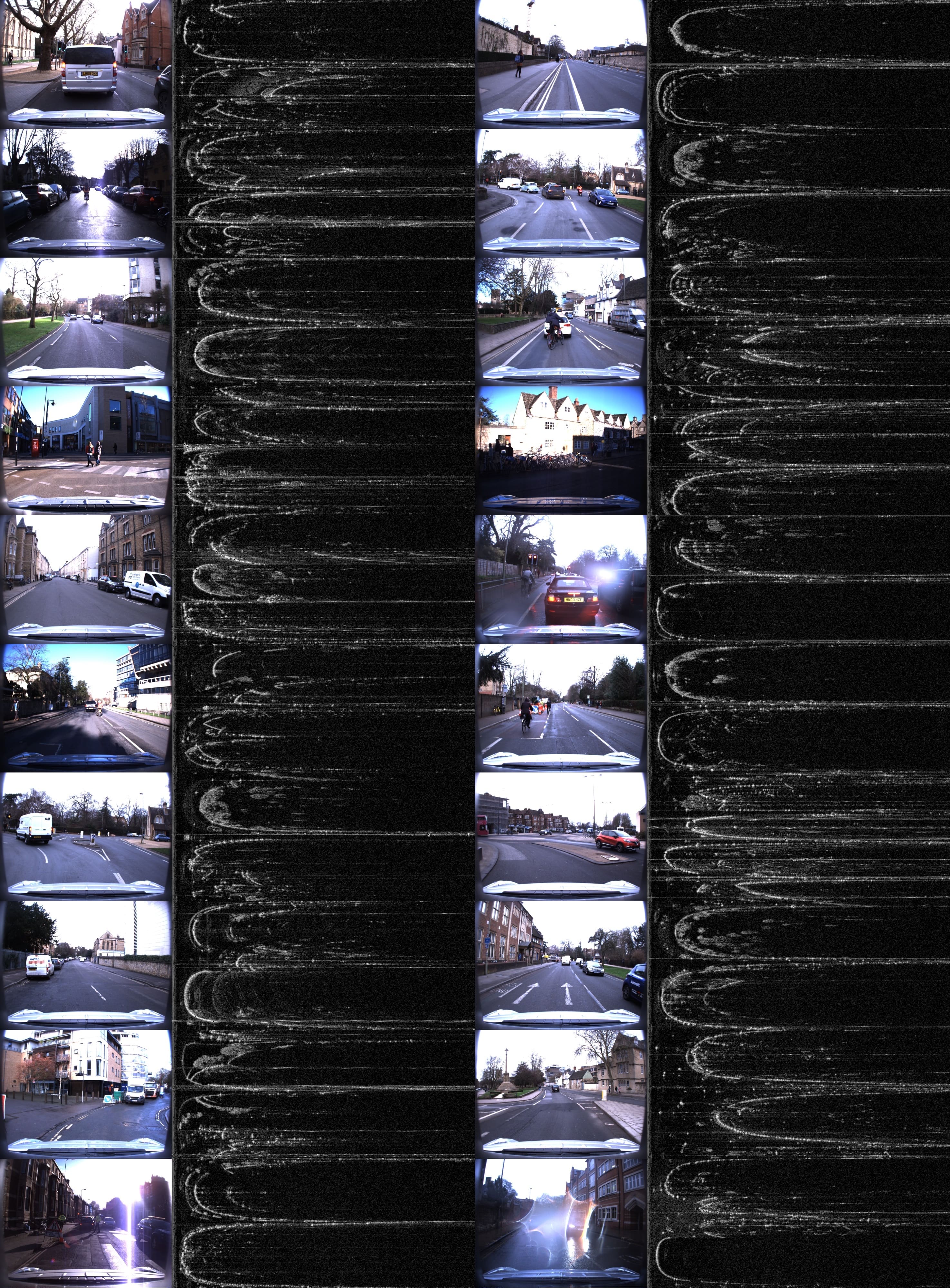}
    \caption{Random pairs of Bumblebee XB3 images (left) with the temporally closest Navtech CTS350-X radar scan (right) from the Oxford Radar RobotCar Dataset, showing the challenging diversity of weather, lighting, and traffic conditions encountered during the period of data collection in Oxford, UK in January 2019.}
    \label{fig:dataset-montage}
\end{figure*}
\fi

This dataset release follows the original Oxford RobotCar Dataset route in Oxford, UK and consists of \SI{32}{} traversals in different traffic, weather, and lighting conditions in January 2019 totalling \SI{280}{\kilo\metre} of urban driving. 
The vehicle was driven manually throughout the period of data collection; no autonomous capabilities were used.
The total download size of the dataset is \SI{4.7}{\tera\byte}.
\ificra
\else
\cref{fig:dataset-montage} shows a random selection of images taken from the dataset, illustrating the variety of situations encountered. 
\fi
\cref{tab:raw-data-summary} lists summary statistics for the raw data collected through the entire month-long collection while \cref{tab:processed-data-summary} lists summary statistics for processed data which are also made available for download.

Every effort was made to follow the exact route for every traversal. However, this was not always possible and slight diversions were made infrequently. 
Additionally, two partial traversals are included which do not cover the entire route.
The GPS/INS data can be used to identify diversions.
However, similarly to \cite{RobotcarDatasetIJRR}, the accuracy of the fused INS solution varied significantly during the course of data collection.
Instead, we suggest using the optimised radar odometry shown in~\cref{fig:dataset-optimised-odometry-data} and discussed in \cref{sec:gt-radar-odometry} as the best available solution of the underlying motion of the radar.

\begin{table}[]
    \centering
    \begin{tabular}{| C{2.75cm} | C{1.71cm} | C{1.2cm} | C{1.175cm} |}
        \hline \textbf{Sensor} & \textbf{Type} & \textbf{Count} & \textbf{Size} \\ \hline
        Bumblebee XB3 & Image & 2,887,776 & 2.2 TB \\ \hline
        Grasshopper 2 & Image & 2,963,601 & 1.6 TB \\ \hline
        LMS-151 & 2D Scan & 5,988,123 & 67.3 GB \\ \hline
        SPAN-CPT GPS & 3D Position & 300,814 & 35.4 MB \\ \hline
        SPAN-CPT INS & 6DoF Position & 3,008,085 & 491.7 MB \\ \hline
        Navtech CTS350-X & Radar Scan & 240,088 & 106.1 GB \\ \hline
        Velodyne Raw & 3D Scan & 2,405,785 & 91.0 GB \\ \hline
    \end{tabular}
    \caption{Summary statistics for collected data.}
    \label{tab:raw-data-summary}
\end{table}

\begin{table}[]
    \centering
    \begin{tabular}{| C{2.75cm} | C{1.71cm} | C{1.2cm} | C{1.175cm} |}
        \hline \textbf{Sensor}              & \textbf{Type} & \textbf{Count}    & \textbf{Size} \\ \hline
        Stereo \gls{vo}       & 6DoF Position & 961,487           & 89.0 MB \\ \hline
        GT Radar Odometry         & 3DoF Position & 240,024           & 28.6 MB \\ \hline
        Velodyne Binary     & 3D Scan       & 2,405,785         & 774.3 GB \\ \hline
    \end{tabular}
    \caption{Summary statistics for processed data.}
    \label{tab:processed-data-summary}
    \ificra
    \else
    \vspace{-2\baselineskip}
    \fi
\end{table}

\subsection{Sensor Calibration}

We include in this release a full set of extrinsic calibration data needed to utilise the additional Navtech and Velodyne sensors while the intrinsics and extrinsics of the sensors from \cite{RobotcarDatasetIJRR} remain unchanged.
\cref{fig:robotcar-platform} illustrates the extrinsic configuration of sensors on the Radar RobotCar platform.
The new LIDAR and radar sensors' extrinsics were calibrated by manually taking measurements of the as-built positions of the sensors as a seed and then performing pose optimisation to minimise the error between laser and radar co-observations.
Precise extrinsic calibrations for each sensor are included in the development tools to be discussed in \cref{sec:sdk}.
As per~\cite{RobotcarDatasetIJRR} the sensor extrinsics are not guaranteed to have remained constant throughout the lifetime of the vehicle.
However, given the relatively short duration of this trial, little degradation is expected.
Given the large overlap in observable environment and diversity of sensor modalities, this dataset provides an excellent test-bed for work on cross-modality calibration and we encourage using our estimates as initial seeds for further research.

\begin{figure}
    \centering
    \includegraphics[width=\columnwidth]{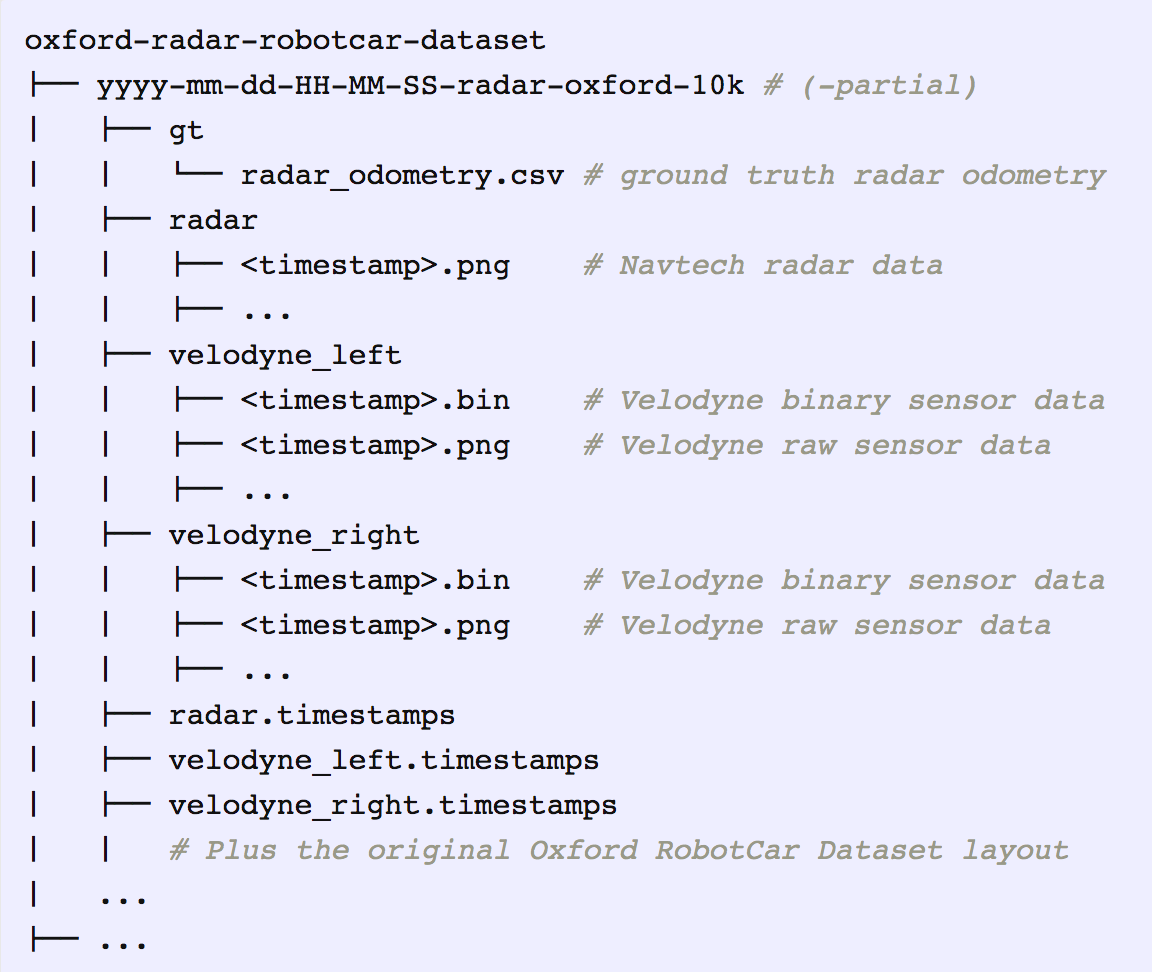}
    \caption{
    Directory layout for the Oxford Radar RobotCar Dataset. 
    When downloading multiple zip archives from multiple traversals, extracting them all in the same directory will preserve the folder structure shown here.
    }
    \label{fig:dataset-directory-structure}
\end{figure}

\subsection{Data Formats}

\cref{fig:dataset-directory-structure} shows the typical directory structure for a single dataset.
In contrast to \cite{RobotcarDatasetIJRR} we do not chunk sensor data into smaller files.
Therefore each zip file download corresponds to the complete sensor data for one dataset traversal (or processed sensor output such as stereo \gls{vo}) with the folder structure inside the archive illustrated in \cref{fig:dataset-directory-structure}. 
The formats for each data type are as follows:

\subsubsection{Radar scans}

are stored as lossless-compressed PNG files in polar form with each row representing the sensor reading at each azimuth and each column representing the raw power return at a particular range.
The files are structured as \verb|<dataset>/radar/<timestamp>.png| where \verb|<timestamp>| is the starting UNIX timestamp of the capture, measured in microseconds.
In the configuration used there are \SI{400}{} azimuths per sweep (rows) and \SI{3768}{} range bins (columns). 

To give users all the raw data they could need we also embed the following \emph{per azimuth} metadata into the PNG image within the first \SI{11}{} columns as follows: 
\begin{itemize}
    \item UNIX timestamp as \verb|int64| in cols 1-8.
    \item Sweep counter as \verb|uint16| in cols 9-10; converted to angle in radians with:
    \begin{lstlisting}
 angle = sweep_counter / 2800 * $\pi$
    \end{lstlisting}
    \item Finally, a \emph{valid} flag as \verb|uint8| in col 11.
\end{itemize}

The \emph{valid} flag is included as there are a very small number of data packets carrying azimuth returns that are infrequently dropped.
To this end, in order to simplify usage for users, we have interpolated adjacent returns so that each provided radar scan has \SI{400}{} azimuths (rows).
If this is not desirable it is advised to simply drop any row which has the \emph{valid} flag set to zero.

\subsubsection{3D Velodyne LIDAR scans}

are provided in two formats, a raw form which encapsulates all the raw data recorded from the sensor for users to do with as they please, or in binary form representing the non-motion compensated pointcloud for a particular scan.

\emph{Raw scans}: are released as lossless PNG files with each column representing the sensor reading at each azimuth.
The files are structured \verb|<dataset>/<laser>/<timestamp>.png|, where \verb|<laser>| is \verb|velodyne_left| or \verb|velodyne_right| and \verb|<timestamp>| is the starting UNIX timestamp of the capture, measured in microseconds.
To give users all the raw data they could need we embed \emph{per azimuth} metadata into the PNG within the following rows: 
\begin{itemize}
    \item Raw intensities for each laser as \verb|uint8| in rows 1-32.
    \item Raw ranges for each laser as \verb|uint16| in rows 33-96, converted to metres with:
    \begin{lstlisting}
 ranges (metres) = ranges_raw * 0.02
    \end{lstlisting}
    \item Sweep counter as \verb|uint16| in rows 97-98; converted to angle in radians with:
    \begin{lstlisting}
 angle = sweep_counter / 18000 * $\pi$
    \end{lstlisting}
    \item Finally, \emph{approximate} UNIX timestamps as \verb|int64| in rows 99-106
\end{itemize}

Timestamps are received for each data packet from the Velodyne LIDAR which includes \SI{12}{} sets of readings for all \SI{32}{} lasers.
We have linearly interpolated timestamps at each azimuth reading.
However, the original received timestamps can be extracted by simply taking every twelfth timestamp.

\emph{Binary scans}: are released as single-precision floating point values packed into a binary file representing the non-motion compensated pointcloud generated from the corresponding raw scan, similar to the Velodyne scan format in~\cite{geiger2013vision}.
The files are structured as \verb|<dataset>/<laser>/<timestamp>.bin|, where \verb|<laser>| is \verb|velodyne_left| or \verb|velodyne_right| and \verb|<timestamp>| is the starting UNIX timestamp of the capture, measured in microseconds. Each scan consists of $(x, y, z, I)$ x $N$ values, where $x$, $y$, $z$ are the 3D Cartesian coordinates of the LIDAR return relative to the sensor (in metres), and $I$ is the measured intensity value.

\subsubsection{Ground Truth Radar Odometry}
\label{sec:data:gt-radar-odometry}

The files \verb|<dataset>/gt/radar_odometry.csv| contain the $SE(2)$ relative pose solution as detailed in \cref{sec:gt-radar-odometry}, consisting of the source and destination frame UNIX timestamps (chosen to be in the middle of the corresponding radar scans), the six-vector Euler parameterisation ($x$, $y$, $z$, $\alpha$, $\beta$, $\gamma$) of the $SE(3)$ relative pose relating the two frames (where $z$, $\alpha$, $\beta$ are all zero but included for compatibility with other pose sources, most notably in the original SDK) and the \emph{starting} source and destination frame UNIX timestamps of the corresponding radar scans which can be used as the \verb|<timestamp>| to load the corresponding radar scan files.

%%%%%%%%%%%%%%%%%%%%%%%%%%%%%%%%%%%%%%%%%%%%%%%%%%%%%%%%%%%%%%%%%%%%%%%%%%%%%%%%
%%% GROUND TRUTH RADAR ODOMETRY
%%%%%%%%%%%%%%%%%%%%%%%%%%%%%%%%%%%%%%%%%%%%%%%%%%%%%%%%%%%%%%%%%%%%%%%%%%%%%%%%

\begin{figure}
    \centering
    \includegraphics[width=\columnwidth,  trim={0, 2.1cm, 0, 2.4cm}, clip]{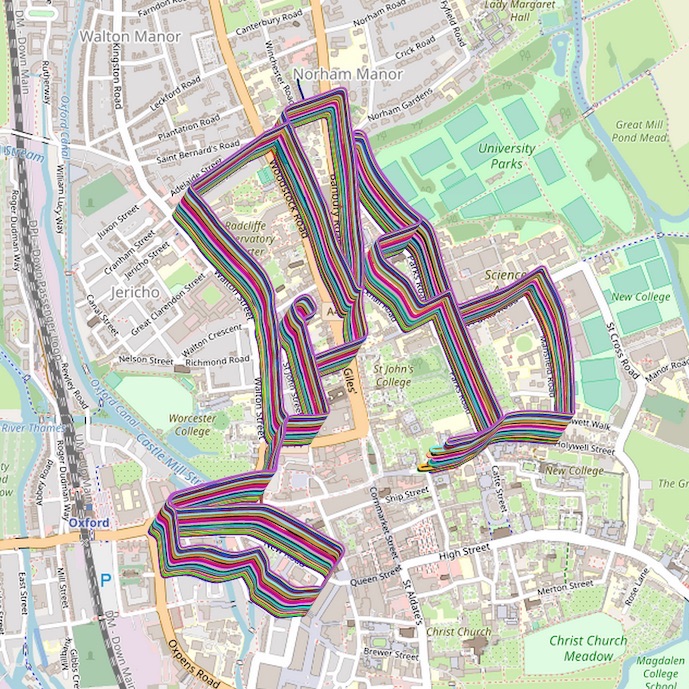}
    \caption{
    Optimised radar odometry plotted on OpenStreetMap~\cite{OpenStreetMap} for each of the \SI{32}{} dataset traversals, where each run is offset for visualisation purposes.
    The trajectories were generated by optimising robust \gls{vo}~\cite{barnes2018driven}, visual loop closures~\cite{cummins2008fab}, and GPS/INS as constraints.
    Map data copyrighted OpenStreetMap contributors and available from \rurl{openstreetmap.org}.}
    \label{fig:dataset-optimised-odometry-data}
    \vspace{-.27cm}
\end{figure}

\section{Ground Truth Radar Odometry}
\label{sec:gt-radar-odometry}

Alongside this dataset we provide ground truth $SE(2)$ radar odometry temporally aligned to the radar data to help further research using this modality for motion estimation, map building, and localisation. 
The poses were generated by performing a large-scale optimisation with Ceres Solver~\cite{ceres-solver} incorporating \gls{vo}, visual loop closures, and GPS/INS constraints with the resulting trajectories shown in \cref{fig:dataset-optimised-odometry-data}.

Specifically, we include all \SI{32}{} dataset traversals and calculate robust \gls{vo} using the approach proposed in~\cite{barnes2018driven}, in which each image is masked with a neural network before generating odometry estimates using~\cite{churchill2012experience}.
Visual loop closures are then found within and across each traversal using FAB-MAP~\cite{cummins2008fab}.
For each traversal we optimise the \gls{vo}, GPS/INS, and individual loop closures in the radar frame to obtain an approximately accurate global $SE(2)$ pose estimate.
Finally, all \SI{32}{} pose chains are jointly optimised with all constraints before interpolating to create the ground truth, time-synchronised radar odometry.

%%%%%%%%%%%%%%%%%%%%%%%%%%%%%%%%%%%%%%%%%%%%%%%%%%%%%%%%%%%%%%%%%%%%%%%%%%%%%%%%
%%% DEVELOPMENT TOOLS
%%%%%%%%%%%%%%%%%%%%%%%%%%%%%%%%%%%%%%%%%%%%%%%%%%%%%%%%%%%%%%%%%%%%%%%%%%%%%%%%

\begin{figure*}
    \centering
    \includegraphics[width=\textwidth]{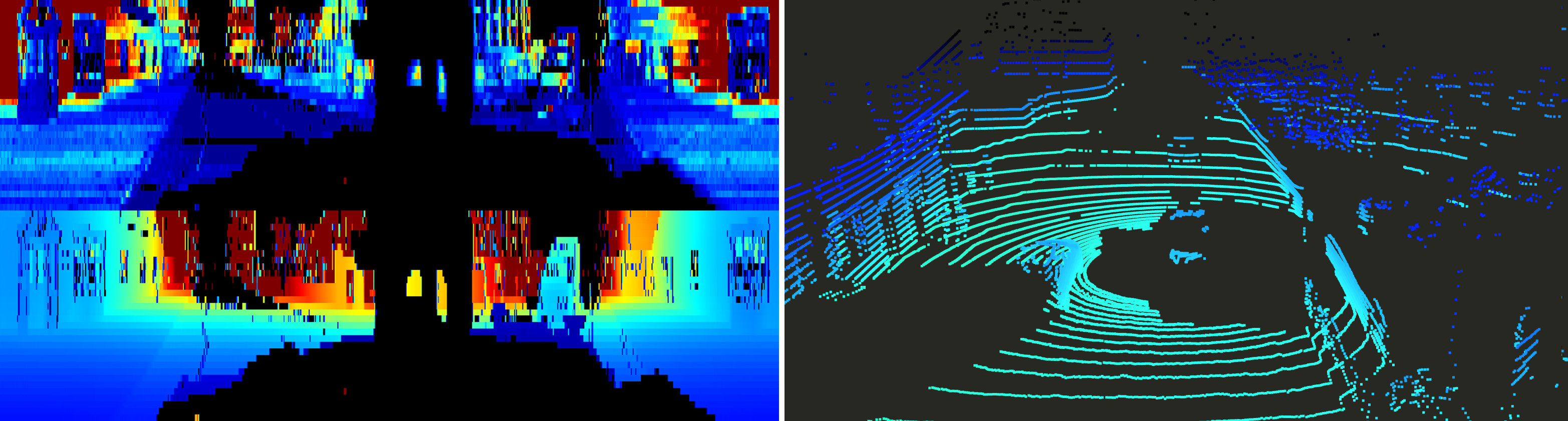}
    \vspace{-1.1\baselineskip}
    \caption{
    Example sensor data from the Velodyne HDL-32E 3D LIDAR. A raw Velodyne scan (left) stores intensities (top) and ranges (bottom) for each azimuth (columns) whereas a binary scan stores the Cartesian pointcloud (right). Tools required to parse the data and perform the raw-to-pointcloud conversion are provided in the \gls{sdk} mentioned in \cref{sec:sdk}. Here the raw scan (left) is shown with invalid pixels set to black and stretched colourmap to improve visibility for the reader.
    }
    \vspace{-.9\baselineskip}
    \label{fig:velodyne-data}
\end{figure*}

\section{Development Tools}
\label{sec:sdk}

We provide a set of MATLAB and Python development tools for easy access to and manipulation of the newly provided data formats; where tools for sensors from the original dataset, such as for imagery, remain unchanged. 
The new tools include simple functions to load and display radar and Velodyne scans as well as more complex functionality such as converting the polar radar data into Cartesian form and converting raw Velodyne data into a pointcloud.
To simplify usage these tools have been merged back into the original Oxford RobotCar Dataset \gls{sdk}\footnote{\rurl{github.com/ori-mrg/robotcar-dataset-sdk}}.
We also provide, and plan to extend, additional functionality useful to the community such as a batch downloader script for this dataset and deep learning data loaders; for up to date information on these please refer to the dataset website.

\vspace{-.1125\baselineskip}
\subsection{Radar Loading and Conversion to Cartesian}

The MATLAB and Python functions \verb|LoadRadar.m| and \verb|load_radar| read a raw radar scan from a specified directory and at a specified timestamp, and return the per-azimuth UNIX timestamps (\SI{}{\micro\second}), azimuth angles (\SI{}{\radian}), and power returns (\SI{}{\decibel}) as well as the range resolution (\SI{}{\centi\metre}) as described previously.
For this data release radar resolution will always equal \SI{4.38}{\centi\metre}.
    
The functions \verb|RadarPolarToCartesian.m| and \verb|radar_polar_to_cartesian| take the azimuth angles (\SI{}{\radian}), power returns (\SI{}{\decibel}) and radar range resolution (\SI{}{\centi\metre}) from a decoded radar scan and converts the polar scan into Cartesian form according to a desired Cartesian resolution (\SI{}{\metre}) and Cartesian size (\SI{}{\pixel}). 

The scripts \verb|PlayRadar.m| and \verb|play_radar.py| produce an animation of the available radar scans from a dataset directory as well as performing polar-to-Cartesian conversion as shown in \cref{fig:radar-data}; please consult this script and the individual functions for demo usage.

\vspace{-.1125\baselineskip}
\subsection{Velodyne Loading and Conversion to Pointcloud}

Similarly, the MATLAB and Python functions \verb|LoadVelodyneRaw.m| and \verb|load_velodyne_raw| read a raw Velodyne scan from a specified directory and at a specified timestamp, of the form \verb|<timestamp>.png|, and return ranges (\SI{}{\metre}), intensities (\SI{}{\uint}), azimuth angles (\SI{}{\radian}), and approximate timestamps (\SI{}{\micro\second}) as described previously. 

The functions \verb|VelodyneRawToPointcloud.m| and \verb|velodyne_raw_to_pointcloud| take the ranges (\SI{}{\metre}), intensities (\SI{}{\uint}), and azimuth angles (\SI{}{\radian}) from a decoded raw Velodyne scan and produce a pointcloud in Cartesian form including per-point intensity values.

The functions \verb|LoadVelodyneBinary.m| and \verb|load_velodyne_binary| read a binary Velodyne scan from a specified directory and at a specified timestamp, of the form \verb|<timestamp>.bin|, and returns a pointcloud in Cartesian form including per-point intensity values.

Finally, the scripts \verb|PlayVelodyne.m| and \verb|play_velodyne.py| produce an animation of the available Velodyne scans from a dataset directory, as shown in \cref{fig:velodyne-data}; please consult this script and the individual functions for demo usage.

%%%%%%%%%%%%%%%%%%%%%%%%%%%%%%%%%%%%%%%%%%%%%%%%%%%%%%%%%%%%%%%%%%%%%%%%%%%%%%%%
%%% SUMMARY AND FUTURE WORK
%%%%%%%%%%%%%%%%%%%%%%%%%%%%%%%%%%%%%%%%%%%%%%%%%%%%%%%%%%%%%%%%%%%%%%%%%%%%%%%%
\section{Summary and Future Work}

We have presented the \textit{The Oxford Radar RobotCar Dataset}, a new large-scale dataset focused on further exploitation of millimetre-wave \gls{fmcw} scanning radar sensors for large-scale and long-term vehicle autonomy and mobile robotics. 
Although this modality has received relatively little attention in this context, we anticipate that this release will help foster discussion of its uses and encourage new and interesting areas of research not previously possible.

In the future, we would like to continue to collect and share large-scale radar datasets in new and challenging conditions and more precisely fine-tune the current extrinsic calibration parameters, perhaps by using publicly available toolboxes designed for radar-LIDAR-camera systems such as~\cite{ICRA19_Domhof}. 
Finally, we would like to investigate semantic scene understanding in radar, perhaps with additionally collecting doppler data, to show that it is a viable alternative for otherwise commonly used sensors like vision and LIDAR. 

%%%%%%%%%%%%%%%%%%%%%%%%%%%%%%%%%%%%%%%%%%%%%%%%%%%%%%%%%%%%%%%%%%%%%%%%%%%%%%%%
%%% ACKNOWLEDGEMENTS
%%%%%%%%%%%%%%%%%%%%%%%%%%%%%%%%%%%%%%%%%%%%%%%%%%%%%%%%%%%%%%%%%%%%%%%%%%%%%%%%
\section{Acknowledgements}

The authors thank all the members of the \gls{ori} who performed scheduled driving over the data collection period.
We would also like to thank our partners at Navtech Radar, without whom this dataset release would not have been possible.

Dan Barnes is supported by the UK EPSRC Doctoral Training Partnership.
Matthew Gadd is supported by Innovate UK under CAV2 -- Stream 1 CRD (DRIVEN).
Paul Newman and Ingmar Posner are supported by EPSRC Programme Grant EP/M019918/1.

%%%%%%%%%%%%%%%%%%%%%%%%%%%%%%%%%%%%%%%%%%%%%%%%%%%%%%%%%%%%%%%%%%%%%%%%%%%%%%%%
%%% BIB
%%%%%%%%%%%%%%%%%%%%%%%%%%%%%%%%%%%%%%%%%%%%%%%%%%%%%%%%%%%%%%%%%%%%%%%%%%%%%%%%

\bibliographystyle{IEEEtran}
\bibliography{biblio}

\end{document}